\pdfoutput=1
\documentclass[11pt]{article}

\usepackage{acl}

\usepackage{times}
\usepackage{latexsym}
\usepackage{CJKutf8}

\usepackage[T1]{fontenc}
\usepackage[utf8]{inputenc}

\usepackage{microtype}

\usepackage{graphicx}
\usepackage{tabularx}
\usepackage{booktabs}
\usepackage{microtype}
\usepackage{arydshln}
\usepackage{footnote}
\usepackage{tablefootnote}
\usepackage{cleveref}
\usepackage{xspace}
\makesavenoteenv{tabular}
\makesavenoteenv{tabularx}
\makesavenoteenv{table*}

\newcommand{\ignore}[1]{}
\newcommand{\cbertbias}{$Bias^{BERT_C}$\xspace}
\newcommand{\wbertbias}{$Bias^{BERT_W}$\xspace}

\newcommand{\wrobertabias}{$Bias^{RoBERTa_W}$\xspace}
\newcommand{\bert}{\textsc{BERT}\xspace}

\newcommand{\amico}{\textsc{AM$^2$iCo}\xspace}

\title{Measuring Context-Word Biases in Lexical Semantic Datasets}

\author{Qianchu Liu,~Diana McCarthy,   ~Anna Korhonen \smallskip \\
Language Technology Lab, TAL, University of Cambridge, UK  \\
\texttt {\{ql261,alk23\}@cam.ac.uk} \\
\texttt {diana@dianamccarthy.co.uk }
}

\begin{document}
\maketitle

\begin{abstract}
 State-of-the-art pretrained contextualized models (PCM) eg. \bert use tasks such as WiC and WSD to evaluate their \textit{word-in-context} representations. This inherently assumes that performance in these tasks reflect how well a model represents the coupled word and context semantics. We question this assumption by presenting the first quantitative analysis on the context-word interaction being tested in major contextual lexical semantic tasks.
%  , taking into account that tasks can be inherently biased and models can rely on different cues than humans. 
 To achieve this, we run probing baselines on masked input, and propose measures to calculate and visualize the degree of context or word biases in existing datasets. The analysis was performed on both models and humans.
%  to decouple biases inherent to the tasks and biases learned from the datasets. 
Our findings demonstrate that models are usually not being tested for word-in-context semantics in the same way as humans are in these tasks, which helps us better understand the model-human gap. Specifically, to PCMs, most existing datasets fall into the extreme ends (the retrieval-based tasks exhibit strong target word bias 
% with the model achieving the best performance using target words alone in the medical domain (eg. COMETA), 
while WiC-style tasks and WSD show strong context bias);
% (2) \amico and Sense Retrieval show less extreme model biases and challenge a model more to represent both the context and target words. 
%  while \amico and Sense Retrieval show lower overall biases and challenge a model more to represent both the context and target words. 
In comparison, humans are less biased and achieve much better performance when both word and context are available than with masked input. 
% This study demonstrates that with heavy context or target word biases, models are usually not being tested for word-in-context representations as humans would behave in these tasks.  
We recommend our framework for understanding and controlling these biases for model interpretation and future task design.
\end{abstract}

\section{Introduction}

% \vspace{-3mm}

\begin{figure}[ht!]
\centering

\includegraphics[width=0.5\textwidth]{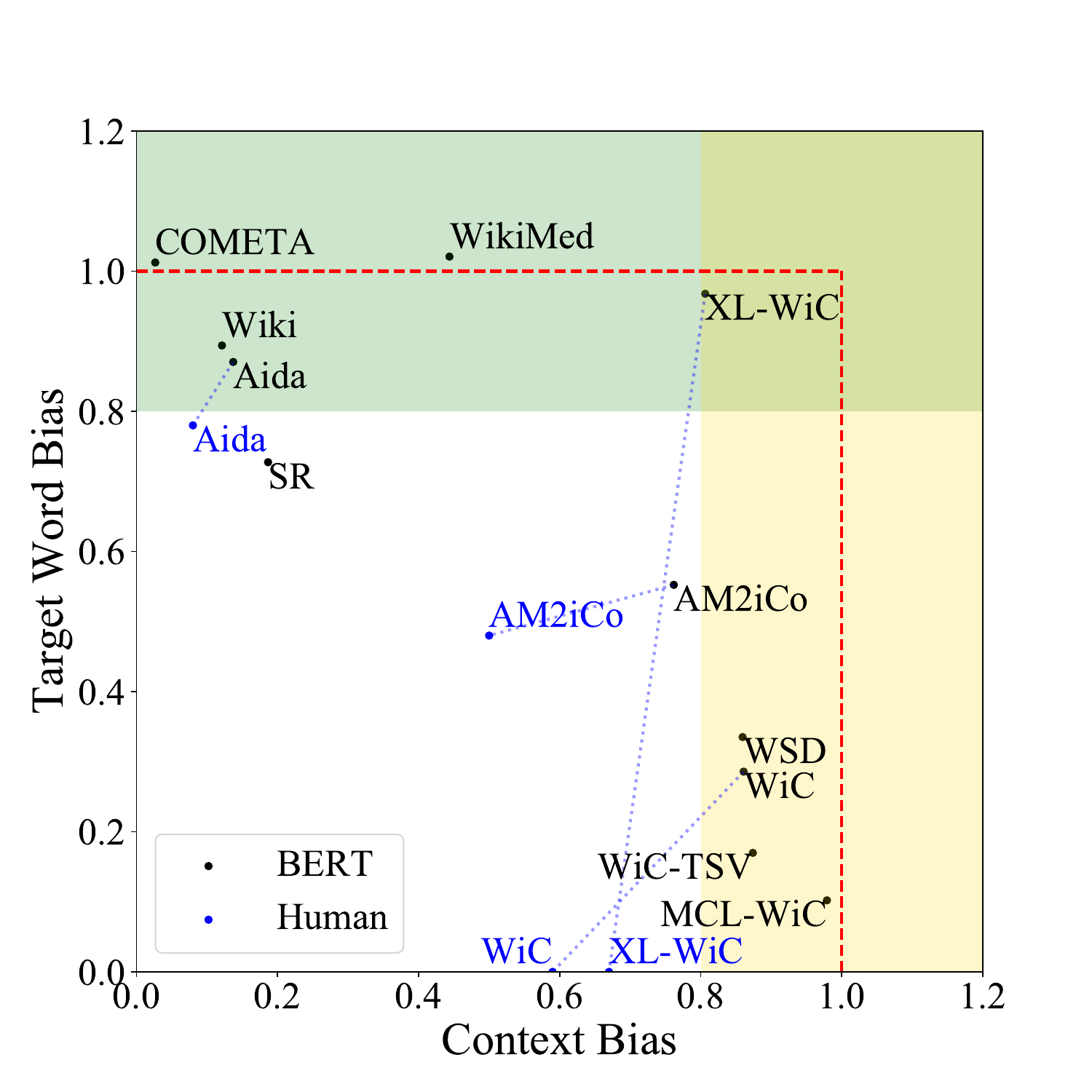}
\caption{Plotting context and target word biases from \textsc{BERT} (black) and humans (blue) across popular context-aware lexical semantic datasets. The green shade and the yellow shade roughly indicate the areas for high target word bias and high context bias (>0.8). We would ideally want a dataset to lie towards the bottom left corner which is bias-free. The dashed red lines indicate 1.0 context (right) and 1.0 target word bias (top), implying a dataset is in effect dealt with by relying on context alone or target words alone.  
 \label{fig:main}}
\end{figure}

Meaning contextualization (i.e., identifying the correct meaning of a target word in linguistic context) is essential for understanding natural language, and has been the focus in many lexical semantic tasks. Pretrained contextualized models (PCMs) have brought large improvements in these tasks including WSD \cite{hadiwinoto-etal-2019-improved,loureiro-jorge-2019-language,huang-etal-2019-glossbert,blevins-zettlemoyer-2020-moving}, WiC \cite{pilehvar-camacho-collados-2019-wic,gari-soler-etal-2019-word} and entity linking (EL) \cite{wu-etal-2020-scalable,broscheit-2019-investigating}. 

These superior performances have been taken as proof that PCMs can successfully model \textit{word-in-context} semantics. Many studies have investigated the process of lexical contextualization in these PCMs. Specifically, \citet{vulic-etal-2020-probing, aina-etal-2019-putting} found language models `contextualize' words in higher layers while the type-level information is better kept in lower layers\footnote{In this study, we do not perform layer-wise analysis as we fine-tune the PCMs to achieve the optimal performance in each task and we expect the relevant information is already surfaced to the last layer \citep{peters-etal-2019-tune}}. \citet{voita-etal-2019-bottom} point out different learning objectives affect the contextualization process, and \citet{gari-soler-apidianaki-2021-lets, pimentel-etal-2020-speakers} show PCMs can capture words' ambiguity levels. 

While most these studies have focused on probing the innerworkings of the PCM feature space, there is no systematic study to quantify the word-context interaction (either learned by PCMs or intrinsic) across different lexical semantic tasks. On one hand, these datasets often vary in their emphasis on context vs target words. For example, we could expect tasks such as WSD and WiC to rely more on context by design as the target words are either given or the same in each input pair \footnote{Notice that the exact amount of context/target word reliance in these tasks is to be tested as humans naturally use both to make prediction.}. On the other hand, models may find shortcuts from datasets to avoid learning the complex word-context interaction. {\bf What is missing in the current literature is an accurate quantification of this word-context interplay being tested in each task so that we can fully understand task goals and model performance.} In particular, we need to flag the situation where a model can solve a task by relying solely on context or the target words. Such heavy word or context reliance hinders a scientific assessment of the models' meaning contextualization abilities as it essentially bypasses the key word-context interaction challenge in 
human understanding of lexical semantics.
% which requires the modeling of both target words and their contexts (Words are frequently ambiguous, but so are contexts. In ``\textit{I like XX .}'', \textit{XX} could have a number of meanings). 
Therefore, we refer to such heavy reliance on target words or context in a contextual lexical semantic dataset as target word biases or context biases \footnote{This is also in line with \citet{gardner-etal-2021-competency}'s claim that all simple feature correlations from partial input are spurious.}.

This study presents an analysis framework to quantify this context-word interaction by measuring context and target word biases across lexical semantic tasks. We first run controlled probing baselines by masking the input to show the context or the target word alone. Based on model's performance on these probing baselines, we calculate two ratios that reflect how much of the model performance in this dataset can be achieved from simply relying on context alone or the target word alone, i.e. the degree of context or target word biases (See Figure~\ref{fig:main} which will be discussed fully in \Cref{sec:results}). The design of the probing baselines follows previous studies that applied input perturbation techniques for model and task analysis in GLUE \cite{pham2020out}, NLI \cite{poliak-etal-2018-hypothesis,wang-etal-2018-glue,talman2021nli} and relation extraction \cite{peng-etal-2020-learning}. While previous probing studies usually assume no meaningful information from corrupted input with no human verification, we provide fairer comparison with model performance by collecting human judgment on the same masked input in four tasks. Such comparison reveals whether the biases are learned by models from the datasets or are inherent in the tasks.

Our key findings are (1) the tasks can be clearly divided into target-word-biased (the retrieval-based tasks), and context-biased (WiC-style tasks and WSD). Among the retrieval-based tasks, domain affects ambiguity level and thus the target word bias: models even achieve the best performance using target words alone in the medical domain. (2) \amico and Sense Retrieval show less extreme model biases and challenge a model more to represent both the context and target words; and (3) a similar trend of biases exists in humans but is much less extreme, as humans find semantic judgment more difficult on masked input and require both word and context to do well in each task. This analysis helps us better understand the nuanced differences between models and humans in existing tasks, and we recommend the framework to be applied when designing new datasets to check whether word and context are required and whether the models rely on the coupled word and context semantics in a similar way to humans.

\section{The Analysis Framework}
\subsection{Task Selection\label{sec:task selection}}
We examine the following contextual lexical semantic tasks, and for illustration, we list example data for each task in Appendix~\ref{sec:task examples}.

\vspace{1.0mm}
\noindent \textbf{Word Sense Disambiguation (WSD).} WSD \cite{navigli2009word,Raganato:2017eacl} requires a model to assign sense labels to target words in context from a set of possible candidates for the target words. Following the standard practice, we use SemCor (sense-annotated texts created from Brown Corpus) as the train set, Semeval2007 as dev, and report accuracy on the concatenated ALL testset. 

\vspace{1.0mm}
\noindent \textbf{The WiC-style Tasks (WiC, WiC-TSV, MCL-WiC and XL-WiC).} To alleviate WSD's requirement for a sense inventory, WiC \citep{pilehvar-camacho-collados-2019-wic} presents a pairwise classification task where each pair consists of two word-in-context instances. The model needs to judge whether the target words in a pair have the same contextual meanings. WiC-TSV \citep{breit-etal-2021-wic} extends the WiC framework to multiple domains and settings. This study adopts the combined setting where each input consists of a word in context instance paired with a definition and a hypernym, and the task is to judge whether the sense intended by the target word in context matches the one described by the definition and is the hyponym of the hypernym. The WiC-style tasks have also been extended to the multilingual and crosslingual settings in MCL-WiC \citep{martelli-etal-2021-mclwic}, XL-WiC \citep{raganato-etal-2020-xl} and more recently in \amico \citep{liu2021am2ico}. MCL-WiC provides test sets for five languages with full gold annotation scores. However, MCL-WiC only covers training data in English. To ensure the analysis will be testing the same data distribution during both training and testing, we will only use the English dataset of MCL-WiC. XL-WiC extends WiC to 12 languages. We perform analysis on the German dataset in XL-WiC (train (50k) and test data (20k)) as it is the only language with sufficient train data and human validation performance. \amico covers 14 datasets, each of which pairs English word-in-context instances with word-in-context instances in a target language. In this study, we perform analysis on the English-Chinese dataset which contain 13k train and 1k test data \footnote{We performed the analysis on other datasets of \amico and found the trend is similar}. 

\vspace{1.0mm}
\noindent \textbf{Sense Retrieval (SR).} With the same train and test data as WSD, SR \cite{loureiro-jorge-2019-language} requires a model to retrieve a correct entry from the full sense inventory of WordNet \cite{miller1998wordnet}.

\vspace{1.0mm}
\noindent \textbf{AIDA and Wikification.}
An important application scenario for testing meaning contextualization is Entity Linking (EL). EL maps a mention (an entity in its context) to a knowledge base (KB) which is usually Wikipedia in the general domain. The target word and its context help solve name variations and lexical ambiguity, which are the main challenges in EL \cite{shen2014entity}. In addition, the context itself can help learn better representations for rare or new entities \cite{schick-schutze-2019-attentive,ji-etal-2017-dynamic}. We test on two popular Wikipedia-based EL benchmarks: AIDA \cite{hoffart-etal-2011-robust} and Wikification (Wiki)  \cite{ratinov-etal-2011-local,bunescu-pasca-2006-using}. AIDA provides manual annotations of entities with Wikipedia and YAGO2 labels for 946, 216 and 231 articles as train, dev and test sets respectively. 
The Wiki Dataset is based on the hyperlinks from Wikipedia. We randomly sampled 50k sentences from Wikipedia as the test and another 50k as the dev set. The rest is used for training. For both AIDA and Wiki, the search space is the full Wikipedia entity list. 

\vspace{1.0mm}
\noindent \textbf{WikiMed and COMETA.}
To test domain effects, we evaluate on two medical EL tasks. We use the WikiMed corpus \cite{vashishth2020improving}, an automatically extracted medical subset 
% containing 358,809 documents 
from Wikipedia, for medical wikification. Each mention is mapped to a Wikipedia page linked to a concept in UMLS \cite{bodenreider2004unified}, a massive medical concept KB. We define the search space as the Wikipedia entities covered in UMLS. With the same Wikipedia ontology but a different domain subset, WikiMed can be directly compared with Wiki for assessing domain influence. 
We also test on COMETA \cite{basaldella-etal-2020-cometa}, a medical EL task in social media. COMETA consists of 20k English biomedical entity mentions from online posts in Reddit. The expert-annotated labels are linked to SNOMED CT \cite{donnelly2006snomed}, another widely-used medical KB. 

We report accuracy for WSD and all the WiC style tasks, and accuracy@1 for retrieval-based tasks including Wiki, AIDA, etc. 

% It has a transformer architecture \cite{vaswani2017attention} pretrained with masked language modeling and next sentence prediction tasks. 
% MLM predicts the vocabulary id of a randomly masked word in a sentence, and NSP trains sentence pair representations to predict the consecutive relation of the pairs. 
% In addition, we offer comparison with a more recent PCM, \textsc{DeBERTa} \cite{he2020deberta}, which improves \textsc{BERT} with two novel techniques: disentangled attention that encodes a word's content and position separately, and an enhanced masked decoder that incorporates absolute position for predicting masked tokens. 
\subsection{Probing Baselines \label{sec:probing baselines}}
% To test how much a model relies on a particular part of the input in a task, we perturb the part of the input during task fine-tuning. \\
% A large drop of performance on this perturbed input implies the bias from the masked part for learning this task. We thus design the following probing baselines that mask different parts of the input to identify these biases. \\
{\bf Context vs. Word:}
For the main experiment, we design the \textsc{Word} baseline where we input only the target word \footnote{In the surveyed tasks, a target word can show different surface variations of number, case and etc. Eg., \textit{breed}, \textit{breeds}.} to the model, and the \textsc{Context} baseline where the target word is replaced with a \texttt{[MASK]} token in the input. The model is then trained and tested on the perturbed input.
A high performance in \textsc{Context} or \textsc{Word} will indicate strong context or target word bias. Example baseline input is shown in Table~\ref{table:wic_exam}. 
{\bf Lower Bound:} Apart from a \textsc{Random} baseline, we also set up a \textsc{Label} baseline where all the input is masked and the learning is only from the label distribution in the task. Notice that training the \textsc{Label} baseline is preferable to simply counting label occurrences in the data as the former can work with both continuous and categorical label space. All the probing baselines are compared with model performance on the full input (\textsc{Full}). We refer to model M's performance in \textsc{Word}, \textsc{Context}, \textsc{Label} and \textsc{Full} as $M_W$, $M_C$, $M_L$ and $M_{Full}$ respectively. {\bf Human Evaluation:} To measure the inherent task biases, we collect human judgment (\textsc{Hum}) for a subset (WiC, XL-WiC, \amico and AIDA) as being representative of the tasks described in \Cref{sec:task selection} and feasible given resources for annotation. WiC, XL-WiC and \amico cover WiC-style datasets in different languages; AIDA is chosen as a representative retrieval-based task. We follow the quality control procedures in \citet{pilehvar-camacho-collados-2019-wic,liu2021am2ico} to recruit two different annotators for each baseline input from \textsc{Context},\textsc{Word} and for \textsc{Full} input in each task. The annotators are recruited from Prolific. They have graduate degrees and are fluent or native in the language of the dataset \footnote{Notice the annotator profiles in this study may be different from the original annotation scheme which was not always clearly specified. Therefore results on \textsc{Full} in this study may be different from figures originally reported. }. In each setup, an annotator is assigned a randomly sampled 100 examples from the test set of each task\footnote{We cannot use the test set for WiC and XL-WiC as the test labels are undisclosed. As the dev set comes from the same distribution of the test, we use dev to estimate human performance in these two tasks. } 
and there is a 50 example overlap between the two annotators for agreement calculation. The annotators are asked to perform meaning judgment in WiC, XL-WiC and \amico, and to find the corresponding Wikipedia pages for entities for AIDA. For \textsc{Context} input where the target words are masked, annotators are encouraged to first guess what the target words could be \footnote{We provide an example annotation guideline in \cref{sec:guideline}. Human has lower but still reasonable agreement in the probing baselines (where there is naturally less information) than with \textsc{Full} input (\cref{sec:agreement}).}. As to the \textsc{Word} input, annotators are asked to think of the most representative meaning of the out-of-context words when performing the tasks. As the pairs of input are always the same word by design in WiC and XL-WiC, we assume humans will give true judgment for all the examples and therefore will score 0.5 on \textsc{Word} input in WiC and XL-WiC. As to human's \textsc{Label} baseline performance, while humans are not given any prior indication of how the task labels will be distributed, it is reasonable to expect that an annotator will give a random choice between the available labels or stick with one label when there is no input. Therefore, we approximate the \textsc{Label} human baseline as being 0.5 for WiC, XL-WiC and \amico, and 0 for AIDA.

\begin{table*}
{\footnotesize
\begin{tabularx}{\textwidth}{p{0.14\textwidth}p{0.42\textwidth}p{0.2\textwidth}p{0.05\textwidth}p{0.05\textwidth}}
\toprule
{\bf Input} & {\bf Sentence1} & {\bf Sentence2}&{\bf \textsc{BERT}} & {\bf \textsc{Human}} \\ 
\midrule
% seen percentage                 & 0.8772                      & 0.7053                    & 0.89346                           & 0.9017                                    & 0.4162  \\ 

\textsc{Full}&Google represents a new [breed] of entrepreneurs .       &The [breed] of tulip .&F&F\\
\textsc{Context}&Google represents a new [MASK] of entrepreneurs .       &The [MASK] of tulip .&F&T \\
\textsc{Word}&breed&breed&T&-\\
\cmidrule(lr){2-5}
\textsc{GuessedWord} &Google represents a new [type] of entrepreneurs .       &The [type] of tulip .&F&T \\
% \midrule

% \textsc{Full} & The pedestrian [misdirected] the out - of - town driver .&[Misdirect] the letter .            & F &F               \\ 
% \textsc{Context}         &  The pedestrian [MASK] the out - of - town driver .  &  [MASK] the letter .              & F & T                \\
% \textsc{GuessedWord}         &   The pedestrian [ignored] the out - of - town driver .      & [Ignore] the letter . &
%  F & T\\
% % \textsc{Advs Context} & [Misdirect] the letter to the out - of - town driver . & The pedestrian [misdirected] the out - of - town driver .&T&F\\
% \midrule
% \textsc{Full}&[Kill] the engine .       &He [kills] the ball . &F&F\\
% \textsc{Context}&[MASK] the engine       &He [MASK] the ball .&F&T \\
% \textsc{GuessedWord} &[Hit] the engine .       &He [hits] the ball .&F&T \\

% \textsc{Advs Context}&Google represents a new [breed] of entrepreneurs .       &Google entrepreneurs presented a new [breed] of tulip . &T&F\\
\bottomrule
\end{tabularx}
\vspace{-1mm}
\caption{Example input of \textsc{Full}, \textsc{Context} and \textsc{Word} in WiC. Target words are in brackets and the original WiC label for the \textsc{Full} example is F. \textsc{GuessedWord} shows human-elicited target words based on \textsc{Context}. Comparing \textsc{Context} and \textsc{GuessedWord} also shows \textsc{BERT}'s contextual bias in WiC as \textsc{BERT} is not sensitive to the target word change.  \label{table:wic_exam}}}
% \textsc{Advs Word} and \textsc{Advs Context} are adversarial examples that flip or preserve the label by changing the target word or the context respectively. ``-'' is not tested.\label{table:wic_exam}}
\end{table*}
\vspace{-1mm}

% \begin{table*}
% % \def\arraystretch{0.8}
% {\small
% \begin{tabularx}{\textwidth}{p{0.16\textwidth}p{0.35\textwidth}p{0.25\textwidth}p{0.05\textwidth}p{0.05\textwidth}}
% \toprule
% {\bf Input} & {\bf Sentence1} & {\bf Sentence2}&{\bf \textsc{BERT}} & {\bf \textsc{Human}} \\ 
% \midrule
% % seen percentage                 & 0.8772                      & 0.7053                    & 0.89346                           & 0.9017                                    & 0.4162  \\ 

% \textsc{Full}&Google represents a new [breed] of entrepreneurs .       &The [breed] of tulip .&F&F\\
% \textsc{Context}&Google represents a new [MASK] of entrepreneurs .       &The [MASK] of tulip .&F&T \\
% \textsc{Word}&breed&breed&T&-\\

% \bottomrule
% \end{tabularx}

% \caption{Examples input of \textsc{Context},\textsc{Word} and \textsc{Full} in WiC.\label{table:wic_exam}}}
% % \textsc{Advs Word} and \textsc{Advs Context} are adversarial examples that flip or preserve the label by changing the target word or the context respectively. ``-'' is not tested.\label{table:wic_exam}}
% \end{table*}

\subsection{Calculating the Bias Measures}
Based on a model $M$'s performance on the full input and on the baseline input, we propose $Bias^{M_C}$ and $Bias^{M_W}$ (as calculated in \Cref{eq:bias_c} and \Cref{eq:bias_w}) to measure the model's context and target word biases in a dataset. $Bias^{M_C}$ is the ratio of $M_C$ to $M_{Full}$ with the \textsc{Label} performance ${M_L}$ deducted from both $M_C$ and $M_{Full}$. $M_L$ has to be deducted as it is unrelated to the input. Otherwise, the ratio will give an inflated bias measurement. $Bias^{M_W}$ is calculated in the same way as $Bias^{M_C}$ except that we replace $M_C$ with $M_W$ in the equation. The two measures can also be seen as $M_C$ and $M_W$ under min-max normalization where the min value is $M_L$ and the max value is $M_{Full}$, and therefore the normalized values can be fairly compared across datasets. $Bias^{M_C}$ and $Bias^{M_W}$ reflect how much of what a model has learned from the input in a dataset can be achieved from context alone or target word alone, which will give us indicators of the degree of context and target word biases in the dataset. These bias indicators will in turn tell us how important the masked part of the input is. For example, we can interpret a $Bias^{M_C}$ of 0.9 as 90\% of what the model has learned from the full input can be achieved from the context alone. The 10\% gap can be gained from adding the masked target word and since this gap is small with a high context bias, we can conclude that the model can do pretty well just from the context alone and it is not learning much from the target word. 

\begin{equation}\label{eq:bias_c}
    Bias^{M_C}=\frac{(M_C-M_L)}{(M_{Full}-M_L)}
\end{equation}

\begin{equation}\label{eq:bias_w}
    Bias^{M_W}=\frac{(M_W-M_L)}{(M_{Full}-M_L)} 
\end{equation}

Like models, humans can also be biased as they can use their prior knowledge (eg. humans can guess the typical meaning of a word without knowing the context) to make predictions based on partial input \citep{gardner-etal-2021-competency}. To measure how much humans can perform on the baseline input will help us understand the biases inherent in a task. We therefore calculate the context and target word bias scores for humans in the same way. 

% as in below (We replace $M$ with $H$ to refer to human performance)

% \begin{equation}\label{eq:bias_c}
%     Bias^{H_C}=\frac{(H_C-H_L)}{(H_{Full}-H_L)}
% \end{equation}

% \begin{equation}\label{eq:bias_w}
%     Bias^{H_W}=\frac{(H_W-H_L)}{(H_{Full}-H_L)} 
% \end{equation}

\subsection{Experiment setup}
The underlying model for our main experiments is \textsc{BERT} \cite{devlin-etal-2019-bert}, the most popular PCM that offers dynamic contextual word representations as bidirectional hidden layers from a transformer architecture. To ensure the general trend of our findings are consistent across different models, we also performed the analysis using \textsc{RoBERTa} \cite{liu2019roberta}, which improves upon BERT by optimized design decisions during training.

We adopt standard model finetuning setups in each task. We use the base uncased variant of \textsc{BERT}\footnote{All PCM configurations are listed in Appendix~D. We also conducted experiments with \textsc{RoBERTa} \cite{liu2019roberta} and reported the results in \Cref{app:roberta}} for general domain experiments and \textsc{PubMedBERT} \cite{pubmedbert} for the medical tasks.
% state-of-the-art methods usually represent labels with glosses in the Wordnet to generalize to unseen labels. 
For WSD, we use \textsc{GlossBert} \cite{huang-etal-2019-glossbert} that learns a sentence-gloss pair classification model based on \textsc{BERT}. 
For the WiC-style tasks, we follow the SuperGlue \cite{wang2019superglue} practices to concatenate \textsc{BERT}'s last layer of \texttt{[CLS]} and the target words' token representations for each input pair, followed by a linear classifier.   
For the retrieval-based tasks including SR and EL, we adopt a bi-encoder architecture to model query and target candidates with BERT \cite{wu-etal-2020-scalable}. For the query, we insert \texttt{[} and \texttt{]} to mark the start and end positions of the target word in context. Each target candidate is reformatted as ``\texttt{[CLS]Name || Description[SEP]}''. \texttt{Name} is an entity title (EL) or synset lemmas from WordNet (SR). \texttt{Description} is the first sentence in an entity's Wikipedia page (Wiki \& WikiMed), a gloss (SR), or n/a (COMETA). The model learns to draw closer the true query-target pairs' representations using triplet loss with triplet miners during fine-tuning \cite{liu2020self}. 
For each experiment, we perform grid search for the learning rate in $[1e-5, 2e-5, 3e-5]$ and select models with early stopping on the dev set. We also run all the models with three random seeds and select the models with the best performance on the dev set. The performance across random seeds are stable as shown by small standard deviations which can be referred to in \Cref{table:dev} in the appendix.

\section{Main Results and Discussion\label{sec:results}}

\begin{figure*}[ht]

\centering
\includegraphics[width=1.0\textwidth]{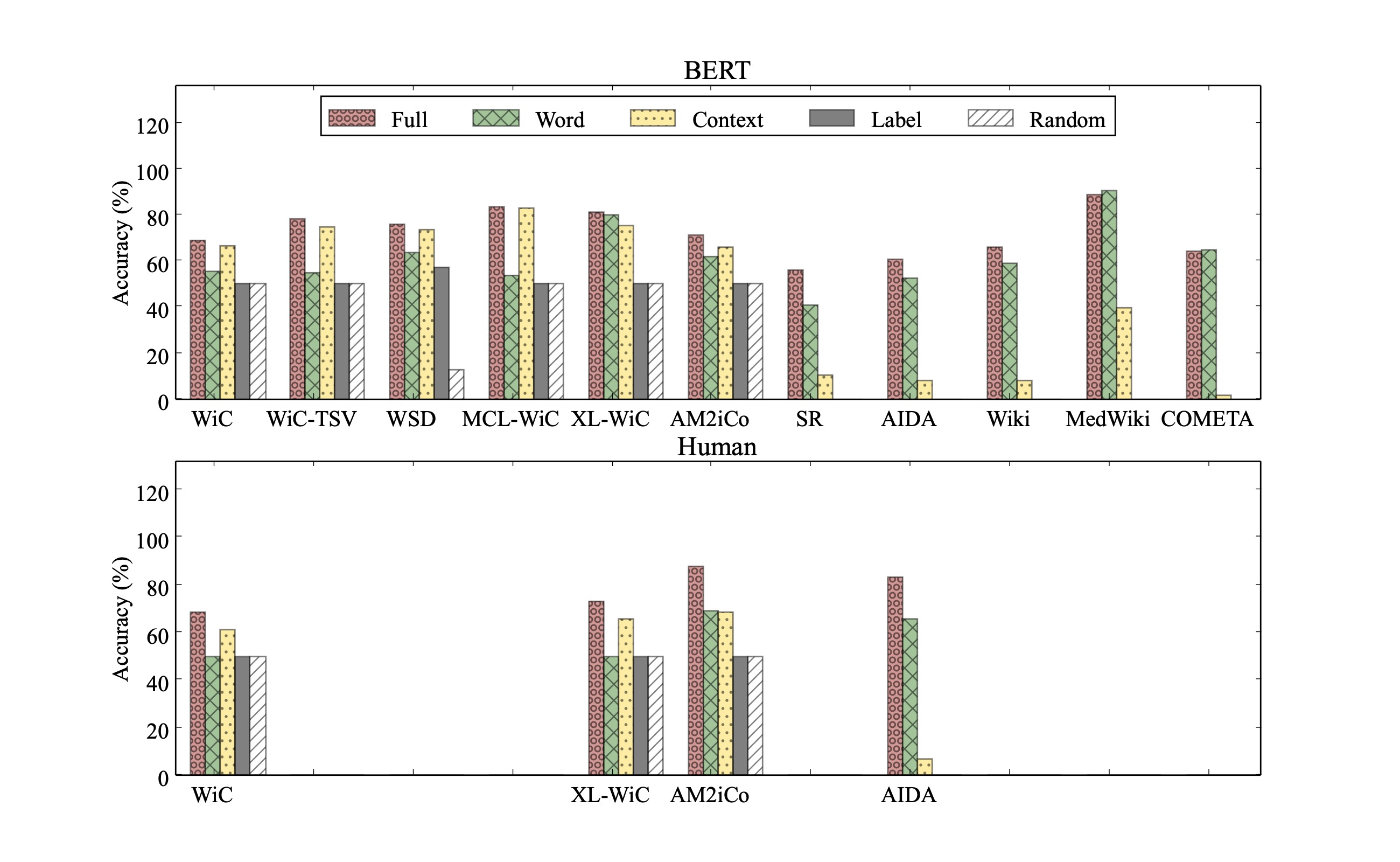}
\vspace{-5mm}
\caption{\textsc{BERT} and human performance on probing baselines across popular context-aware lexical semantic tasks. For the retrieval-based tasks, we report @1 accuracy, and the \textsc{Label} and \textsc{Random} baselines are not visible as they are close to 0. 
 \label{fig:baselines}}
\end{figure*}

We report \textsc{BERT}'s baseline performance in \Cref{fig:baselines}, based on which we calculate \cbertbias and \wbertbias for each dataset and plot the results (black dots) in \Cref{fig:main} (We also report \textsc{RoBERTa} biases in \Cref{app:roberta} and found a similar trend). For comparison, we plot human baseline performance and biases alongside the model performance in each figure. 

\subsection{Model biases}

\noindent \textbf{Models can learn extreme context or target word biases from the datasets.} 
One obvious observation from \Cref{fig:main} is that, probed with \textsc{BERT}, most of the datasets lie close to the dashed red lines: tasks such as WiC and MCL-WiC lie towards the right and are close to the vertical red line which indicates 1.0 context bias; the retrieval-based tasks such as WikiMed and Wiki lie towards the top and are close to or even surpass the horizontal red line which indicates 1.0 target word bias. 
This pattern indicates that \textsc{BERT} can score highly on these datasets by relying only on the target words or only on the context. In other words, context or target words can be much ignored when the model learns to solve the tasks. It is therefore questionable how much word-context interaction, which requires the modeling of both word and context representations, is actually learned by \textsc{BERT} when applied to these tasks.

Moreover, the datasets tend to concentrate in two corners. That is, models usually learn strong bias from either context or the target word: the retrieval-based datasets (eg. Wiki) lie in the top left corner, showing large target word bias and low context bias; the WiC style datasets and WSD lie in the bottom right corner with large context bias and low target word bias. XL-WiC is an exception as it contains both strong context and target word biases. We will come back to this later in \Cref{sec:human vs model} where we compare model and human performance.  

\vspace{1.0mm}
\noindent \textbf{\amico and SR are closest to testing word-context interaction from models.}
There are few existing datasets that in effect require the modeling of the context-word interaction, which should result in both low context and target word biases. SR and \amico can be seen as two such datasets which, in \Cref{fig:main}, can be found further inside of the red lines towards the bias-free left bottom corner. This is because these two tasks are designed to require balanced attention over context and target words.  In SR, a system needs to model the target words in order to retrieve all the possible senses associated with the word, and because there is plenty of ambiguity in the dataset, context is also crucial to identify the correct sense. \amico was specifically designed to include adversarial examples to penalize models that rely only on the context, and therefore elicits the lowest context bias from models among the WiC-style tasks. As such, SR and \amico are the closest tasks that we have to test word-context interaction. 

% The strong target word bias in XL-WiC is somewhat unexpected as the task by design should not leak any label information from the target words as they are always the same in each input pair. 

\vspace{1.0mm}
\noindent \textbf{Domains affect lexical ambiguity and the target word bias.}

\begin{table}[ht]%[t!]
\centering
\scriptsize{

\begin{tabularx}{0.49\textwidth}{p{0.11\textwidth}p{0.03\textwidth}p{0.03\textwidth}p{0.03\textwidth}p{0.05\textwidth}p{0.02\textwidth}}
\toprule 
&SemCor & Wiki & AIDA  & WikiMed & COMETA \\ 
\midrule
 Sense Entropy& .210              &.060         & .044                                           & .026                                     & .001                      \\ 
 \wbertbias & .727 & .894&.871&	1.021&	1.012\\
 \wrobertabias&.732&		.899& .832 & .996 &	1.1780\\
\bottomrule

\end{tabularx}
\caption{Target Word Bias and Sense Entropy across retrieval-based tasks.\label{table:sense_entr}}}
\end{table}

\vspace{1.0mm}
The retrieval-based tasks in this study offer comparison between two domains, general vs medical, by comparing Wiki/AIDA and WikiMed. The target word bias is increased in the medical domain where relying on the target words alone gives the best performance (i.e. COMETA and WikiMed both have > 1.0 target word bias). Such divergence across domains is arguably caused by the different degrees of lexical ambiguity in these tasks. In particular, domain could reduce ambiguity \cite{magnini2002role,koeling-etal-2005-domain}, and therefore affect the importance of the context and therefore the target word bias. As a quantitative measure for lexical ambiguity, we calculate average sense entropy across all words in each task's training data, see Table~\ref{table:sense_entr}. Confirming our hypothesis, sense entropy (lexical ambiguity) in a task does roughly correlate with the model's target word bias: the medical domain tasks (WikiMed and COMETA) contain the lowest lexical ambiguity as reflected by the lowest sense entropy, and therefore missing context in these two tasks will not bring so much negative impact on the model performance, resulting in the highest target word biases; whereas higher sense entropy and thus higher lexical ambiguity (eg. Wiki and then SR) will necessarily require context alongside the target word, which leads to lower target word biases.

\noindent \textbf{Context can harm model performance in Medical EL.}
We notice that the model's target word bias in COMETA and WikiMed can go beyond 1.0, indicating that the model learning is dominated entirely by the target words with the context being useless or even harmful. This comes as a surprise as medical EL has been treated as a contextual lexical semantic task where the context is usually provided in the hope for higher modeling accuracy. We examined the errors from \textsc{Full} as compared with \textsc{Word}, and we found that the model tends to get distracted by related context words. Table~\ref{tab:errana_med} shows an example where the retrieval model selects the entry that is closer to a context word (``Miltonia'') than to the target word (``Miltoniopsis''), but in fact knowing the target word alone in this case is sufficient to retrieve the correct label. This indicates that the model has not learned a good strategy to incorporate word and context representations from the datasets (i.e. not knowing when to focus on the context and when to focus on the target words).

\begin{table}
{\scriptsize
\begin{tabularx}{0.5\textwidth}{p{0.03\textwidth}p{0.13\textwidth}p{0.18\textwidth}p{0.05\textwidth}}
\toprule
    & {\bf Input} & {\bf Retrieved concept entry} & {\bf Result}\\
    \midrule
     \textsc{Full}&Formerly many more species were attributed to ``Miltonia'', ... including [Miltoniopsis] and Oncidium ...& miltonia: miltonia is an orchid genus comprising twelve epiphyte species and eight natural hybrids. & Wrong \\
     \cmidrule(lr){2-4}
    %  \hdashline
     \textsc{Word}&Miltoniopsis & miltoniopsis: miltoniopsis is a genus of orchids native to costa rica and etc. & Correct \\
\bottomrule
\end{tabularx}
\caption{Error analysis on BERT predictions on \textsc{Full} and \textsc{Word} from WikiMEd.\label{tab:errana_med}}}
\end{table}

\subsection{Human vs Model\label{sec:human vs model}}
% The human-model comparison in \Cref{fig:main} and \Cref{fig:baselines} gives us the following findings.

\noindent \textbf{There are inherent task biases.}
Our first finding is that humans show a similar trend of biases in the tasks in comparison to model biases (except for XL-WiC). This is evident from \Cref{fig:main} where, with the human bias indicators, WiC still lies near the bottom right corner with relatively high context bias; AIDA lies near the top left corner with high target word bias and \amico remains in the middle. This confirms that there are some degrees of biases inherent in the task design so that humans can also rely on either target words or context alone to perform the task to some extent. 

\vspace{1.0mm}
\noindent \textbf{Humans are less biased than models.}

\begin{figure}[ht]

\centering
\includegraphics[width=0.5\textwidth]{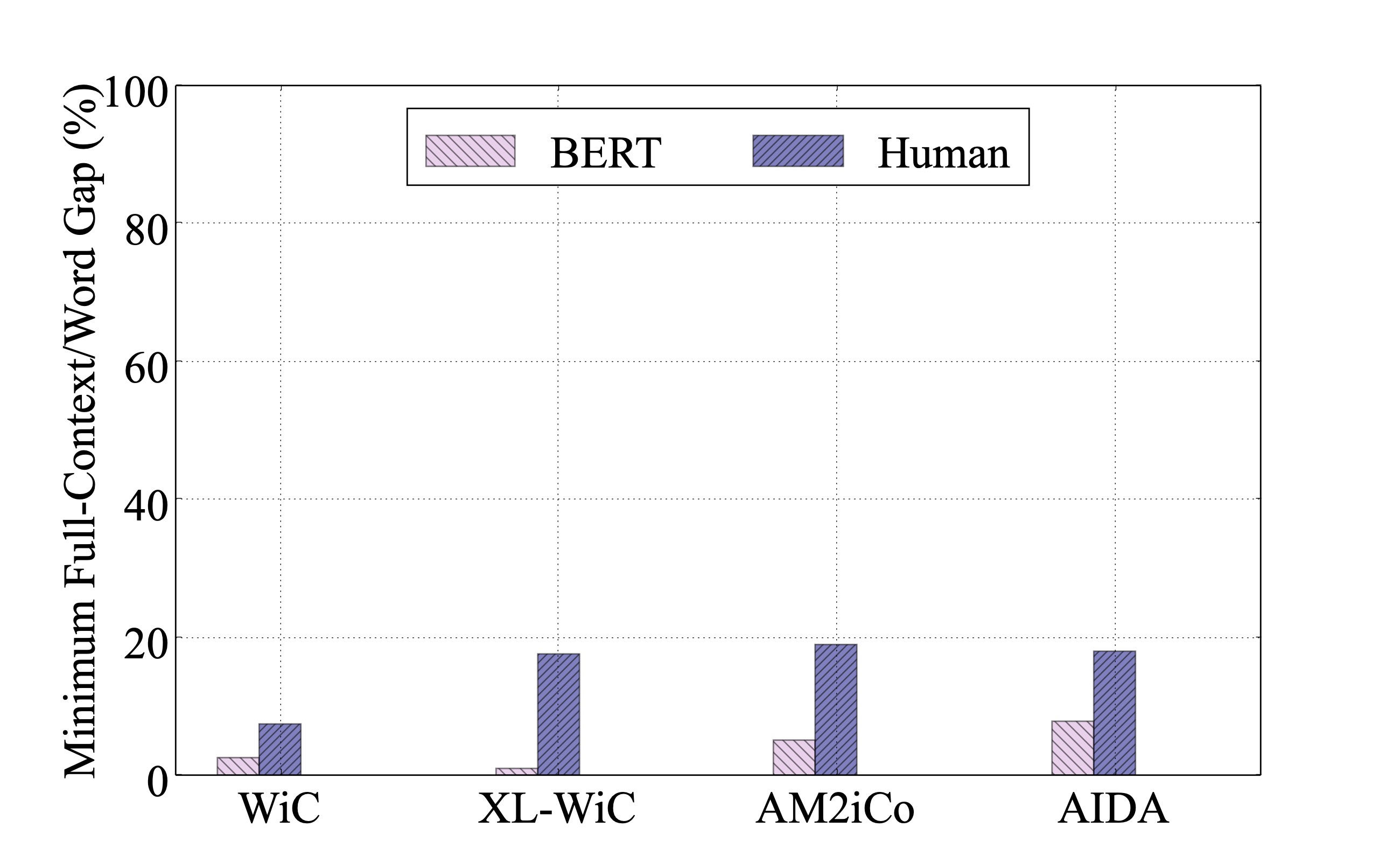}

\caption{The minimum gap between \textsc{Full} and \textsc{Context} or \textsc{Word}, i.e. min(\textsc{Full}-\textsc{Context},\textsc{Full}-\textsc{Word}) with \textsc{BERT} and human performance. A small gap will indicate strong bias.
 \label{fig:baselines_gqp}}
\end{figure}
That being said, the second finding and the more important one is that humans exhibit overall much weaker biases in comparison with models in all the four tasks. If we compare human performance with model performance in \Cref{fig:baselines}, we can see the \textsc{Context} and \textsc{Word} baseline scores are lower in comparison to \textsc{Full} from human performance. For clearer comparison, we calculate and plot the minimum gap between \textsc{Full} to either of the two baselines in \Cref{fig:baselines_gqp}, and we can see substantial difference between humans and models where humans exhibit much larger gaps across the four tasks. The much larger gaps from humans also result in all the four tasks moving further towards the left-bottom ``bias-free'' corner as shown \Cref{fig:main}. In other words, humans are more likely than models to rely on both word and context as the absence of either part will lead to much more negative impact for humans when performing these tasks. 

The most dramatic difference is in XL-WiC where the model's strong target word bias disappears in humans.
The task of XL-WiC by nature should not leak any information from the target word alone (hence 0 target word bias for humans) as the input pair will always contain the same target word.
The high target word bias from models comes from the fact the dataset does not contain sufficient ambiguous cases where the same word pair can have both true and false labels dependent on the contexts. We confirm this by calculating the per-word average label entropy of the training data for all WiC-style tasks \footnote{We disgard words that only occur once} in Table ~\ref{table:label_entr}, and we found XL-WiC has the lowest label entropy  as 0.2084 , and on average a word pair has the same label for 87\% of the examples it appears in the dataset. Therefore, the model learns correlation between the word itself and the label without needing context for disambiguation.

\begin{table}[h]%[t!]
\centering
\scriptsize{
\begin{tabularx}{0.49\textwidth}{p{0.07\textwidth}p{0.03\textwidth}p{0.05\textwidth}p{0.06\textwidth}p{0.07\textwidth}p{0.07\textwidth}}
\toprule 
&WiC&	\amico	&XL-WiC & MCL-WiC&	WiC-TSV \\ 
\midrule
% seen percentage                 & 0.8772                      & 0.7053                    & 0.89346                           & 0.9017                                    & 0.4162                      \\ 
%percentage of unambiguous words & 0.7503                      & 0.926                     & 0.9254                            & 0.959                                     & 0.9992                      \\ 
 LE& 0.4022&	0.5383&	0.2084&	0.2976&	0.4212                      \\ 
\bottomrule

\end{tabularx}
\caption{Label Entropy (LE) across WiC style tasks.\label{table:label_entr}}}
\end{table}

Finally, the fact that models cannot achieve a similarly large jump of performance from masked input to \textsc{Full} like humans could indicate the word-context interaction is particularly challenging for models and this might eventually explain the model-human gap. We take \amico as an example that explicitly requires word-context interaction (Table~\ref{table:amico}). While \bert achieves comparable results with humans in \textsc{Context} and \textsc{Word}, a significantly larger human-model gap is found in \textsc{Full}, indicating the word-context interaction is what the model lacks the most to achieve human-like performance. 

\begin{table}[h]%[t!]
\centering
\scriptsize{
\begin{tabular}{lccc}
\toprule 
& \textsc{Context} & \textsc{Word} & \textsc{Full} \\ 
\midrule
 Human&    69 & 68.5 &   87.9              \\ 
 \bert & 66 & 61 & 71 \\
 Human - \bert& 3 & 7.5 & {\bf 16.9} \\
\bottomrule

\end{tabular}
\caption{Model-human gap in \textsc{Context}, \textsc{Word} and \textsc{Full} in \amico.\label{table:amico}}}
\end{table}

\noindent \textbf{Target words are important in WiC for humans.}
The much lower context bias from humans in tasks such as WiC suggests that the absence of the target words drastically decreases performance. In fact, human \textsc{Context} baseline (0.61) is even worse than \bert (0.65) as shown in \Cref{fig:baselines}. This may also come as a surprise, considering that target words are always the same and only the context is different in each pair of input. We examined human response in \textsc{Context} and found that humans can guess another valid target word based on the context, which gives a different prediction. \Cref{table:wic_exam} shows such an example. While the original WiC label of the input is F, our annotator gave T for the \textsc{Context} input, guessing the target word is \textit{type}. This is a reasonable prediction as \textit{type} fits the contexts and does hold its meaning across the two sentences. We refer to this new example with human-elicited target words as \textsc{GuessedWord} input. The same annotator was able to give the WiC label F when we reveal the original target word (\textit{breed}) which has the specific meaning of \textit{species} in sentence1 and \textit{personality} in sentence2 (see the \textsc{Full} input in Table~\ref{table:wic_exam}). \textsc{BERT} however still predicts F regardless of the target word change in this \textsc{GuessedWord} example. 
 
 As qualitative analysis on the human-model discrepancy on \textsc{Context}, we examined 20 cases where annotators did not predict WiC labels (from the corresponding \textsc{Full} input) while \textsc{BERT} did. In 11 cases, humans guessed other valid target words to justify their predictions. We then perform preliminary analysis to test \textsc{BERT} on all the 11 \textsc{GuessedWord} cases where the human-elicited target words change the labels (We show more examples in \Cref{table:wic_exam_more}), and found that for 7 out of 11, \textsc{BERT} is insensitive to the changed target words and maintains its original prediction. This suggests \textsc{BERT} does not appreciate the same word-context interaction as humans, and is making prediction mainly based on contexts rather than modeling contextual lexical semantics in WiC.

\section{Implications for future dataset design}
We recommend this analysis framework in future dataset design and result interpretation for contextual word representation evaluation. In particular, we recommend (1) creating probing baselines by masking the context and word (if relevant), and (2) providing a sample to humans (details in \cref{sec:probing baselines}), (3) and comparing human and model performance of full input vs the masked baseline/s, and then calculate bias indicators. In terms of task design, we would ideally want both models and humans to show low baseline performance and thus low bias measures. When interpreting the results, apart from evaluating model performance on the \textsc{Full} input, we should also ensure the model shows a human-like gap in performance (between \textsc{Full} and the baseline(s)) on the same data.

\section{Conclusion and Limitations}
This study presented an analysis framework to disentangle and quantify context-word interplay in application of popular contextual lexical semantic benchmarks. With our proposed bias measures, we plot datasets on a continuum, and we found that, to models, most existing datasets lie on the two ends with excessive biases (WiC-style tasks and WSD are heavily context-biased while retrieval-based tasks are heavily target-word-biased) that essentially bypass the key challenges in word-context interaction. SR and \amico have been identified as two tasks that have less extreme biases and therefore can better test the representation of both word and context, and we call for more tasks that challenge models to do so. In addition, we identify that the degree of lexical ambiguity as a byproduct of domain affects target word bias (medical>general) in retrieval-based tasks. Most importantly, we differentiate biases learned by models and task-inherent biases by collecting human responses on the same baseline input. We found that models' heavy context and target word biases are not attested to the same extent in humans who usually need both context and target words to perform well in the tasks. This suggests that models are relying on different cues instead of modeling contextual lexical semantics as intended by the tasks. Our paper highlights the importance of understanding these biases in existing datasets and encourages future dataset and model design to control for these biases and to focus more on testing the challenging word-context interaction in context-sensitive lexical semantics. 

One limitation of this study is that we do not have large-scaled quantitative evidence to pinpoint the cues the models rely on from partial input \footnote{It could be that models like \bert (trained with masked language modelling) are genuinely better than humans in exploiting partial input.}. Possible future directions will be to design such ablation studies to identify any spurious correlations the models have learned and introduce adversarial examples that penalize sole reliance on context or target words in both task design and model training. 

\section*{Acknowledgments}
We thank the anonymous reviewers
for their helpful feedback. We acknowledge Peterhouse College at University of Cambridge for supporting Qianchu Liu's PhD research. The work was also funded by the EPSRC grant EP/T02450X/1 awarded to Anna Korhonen.

\bibliography{anthology,custom}
\bibliographystyle{acl_natbib}

\newpage
\appendix
\section{Task examples \label{sec:task examples}}
Table~\ref{tab:task example} lists example input and labels for tasks surveyed in this study.

\begin{table*}
{\scriptsize
\begin{tabularx}{\textwidth}{p{0.11\textwidth}p{0.25\textwidth}p{0.15\textwidth}p{0.27\textwidth}p{0.08\textwidth}}
\toprule
     {\bf Task} &{\bf Input} & {\bf Label} & {\bf Label Space} & {\bf Metrics}\\
    \midrule
    \textbf{WiC}& Room and [board].      
    
    He nailed [boards] across the windows. & F & T or F & Acc\\
    \cmidrule(lr){2-5}
    {\bf WiC-TSV}& I spent my [spring] holidays in Morocco.     
    
    the season of growth; season, time of the year &   T & T or F & Acc\\
        \cmidrule(lr){2-5}

    \textbf{MCL-WiC}& Bolivia holds a key  [play]  in any process...   
    
    A musical [play]  on the same subject... & F & T or F & Acc\\
        \cmidrule(lr){2-5}

     \textbf{XL-WiC}& Herr [Starke] wollte uns kein Interview geben.  
    
    Das kann ich dir aber sagen: Wenn die Frau [Starke] kommt... & T & T or F & Acc\\
        \cmidrule(lr){2-5}

     \textbf{\amico}& ...\begin{CJK*}{UTF8}{gbsn}航天员训练及[阿波罗]中飞船\end{CJK*}...
     ...the six [Apollo] Moon landings...  & T & T or F & Acc\\
    \cmidrule(lr){2-5}
    {\bf WSD}& The [art] of change-ringing is peculiar to the English...& {\bf art}: a superior skill that you can learn by study and practice and observation& {\bf art}: the creation of beautiful or significant things
    
    {\bf art}: the products of human creativity; works of art collectively
    
    ...(all possible meanings of \textit{art})& F1\\
    \cmidrule(lr){2-5}
    {\bf SR}&The [art] of change-ringing is peculiar to the English...& {\bf art}: a superior skill that you can learn by study and practice and observation& {\bf art}: a superior skill that you can learn by study and practice and observation
    
    {\bf door}: a swinging or sliding barrier that will close the entrance...
    
    ... PLUS all other entries in WordNet & Acc\\
    \cmidrule(lr){2-5}
    
    {\bf Wiki} & an additional [Hash] literal syntax using colons for symbol keys...& \textbf{hash table}: in computing , a hash table ( hash map ) is a data structure... & \textbf{hash table}: in computing , a hash table ( hash map ) is a data structure ...
    
    \textbf{united kingdom}: the United Kingdom of Great Britain and Northern Ireland...
    
    ... (all entries in Wikipedia)& Acc@1\\
    \cmidrule(lr){2-5}
    
     {\bf WikiMed}&The flowers produce pollen, but no nectar. Various bees and flies visit the flowers looking in vain for nectar, for instance [sweat bees] in the genera ``Lasioglossum'' and ``Halictus''...& \textbf{halictidae}: the Halictidae is the second largest family of Apoidea bees. & \textbf{halictidae}: the Halictidae is the second largest family of Apoidea bees.
     
     \textbf{eomecon}: eomecon is a monotypic genus of fl
owering plants in the poppy family...

     ... (all entries in the medical section of Wikipedia)&Acc@1\\
     
     \cmidrule(lr){2-5}
     
     {\bf COMETA}& I am [spacey] because I am thinking and daydreaming about my obsession. & \textbf{dizziness (finding)} & \textbf{dizziness (finding)}
     
     \textbf{large intestine}
     
     ...PLUS all other entries in SNOMED CT&Acc@1
     \\

    %  \hdashline
    
\bottomrule
\end{tabularx}
\caption{Examples for a selection of context-sensitive lexical semantic tasks surveyed in this thesis. Acc: accuracy; $\rho$: Spearman's correlation; $r$: Pearson's correlation; P\&R: precision and recall. \label{tab:task example}}}

\end{table*}
% \section{An example of the target word bias in WikiMed\label{app:wikimed}}
% \begin{table*}
% {\scriptsize
% \begin{tabularx}{\textwidth}{p{0.05\textwidth}p{0.25\textwidth}p{0.5\textwidth}p{0.1\textwidth}}
% \toprule
%     {\bf Baseline}& {\bf Input} & {\bf Retrieved concept entry} & {\bf Result}\\
%     \midrule
%      \textsc{Full}&Formerly many more species were attributed to "Miltonia", ... including [Miltoniopsis] and Oncidium ...& miltonia: miltonia is an orchid genus comprising twelve epiphyte species and eight natural hybrids. & Wrong \\
%     %  \hdashline
%      \textsc{Word}&Miltoniopsis & miltoniopsis: miltoniopsis is a genus of orchids native to costa rica and etc. & Correct \\
% \bottomrule
% \end{tabularx}
% \caption{Error analysis on \textsc{Full} and \textsc{Word} BERT predictions on WikiMEd.\label{tab:errana_med}}}
% \end{table*}

\section{Dev performance}

\Cref{table:dev} shows BERT biases calculated over three runs on the dev set with standard deviation reported. 

\begin{table*}[h]
{\footnotesize
\begin{tabularx}{\textwidth}{lp{0.05\textwidth}p{0.05\textwidth}p{0.05\textwidth}p{0.05\textwidth}p{0.04\textwidth}p{0.06\textwidth}p{0.05\textwidth}p{0.05\textwidth}p{0.05\textwidth}p{0.06\textwidth}p{0.05\textwidth}}
\toprule
 & WiC & WiC-TSV & WSD & MCL-WiC & XL-WiC & \amico & SR & AIDA & Wiki & MedWiki & COMETA\\
\midrule
\wbertbias&0.473&	0.266	&0.346&	0.122&	0.903&	0.665&	0.648&	0.910&	0.946&	1.024&	1.017\\
& (0.016)& (0.043)	& (0.015)	& (0.007)	& (0.002)	& (0.008)	& (0.012)	& (0.007)	& (0.002)	& (0.022)	& (0.034)\\

\cmidrule(lr){2-12}
\cbertbias& 1.055&	0.890&	0.874&	0.864&	0.844&	0.768&	0.237&	0.241&	0.308&	0.447&	0.028\\
& (0.017)	& (0.028)	& (0.020)	& (0.043)	& (0.002)	& (0.016)	& (0.011)	& (0.015)	& (0.003)	& (0.010)	& (0.010)\\
\bottomrule
\end{tabularx}

\caption{Average context and target word biases over three runs with three different random seeds on the dev set in each dataset. Standard deviation is reported in the parenthesis.\label{table:dev}}}
% \textsc{Advs Word} and \textsc{Advs Context} are adversarial examples that flip or preserve the label by changing the target word or the context respectively. ``-'' is not tested.\label{table:wic_exam}}
\end{table*}

\begin{table*}
{\scriptsize
\begin{tabularx}{\textwidth}{p{0.11\textwidth}p{0.32\textwidth}p{0.37\textwidth}p{0.04\textwidth}p{0.04\textwidth}}
\toprule
{\bf Input} & {\bf Sentence1} & {\bf Sentence2}&{\bf \textsc{BERT}} & {\bf \textsc{Hum}} \\ 
\midrule
% seen percentage                 & 0.8772                      & 0.7053                    & 0.89346                           & 0.9017                                    & 0.4162  \\ 

\textsc{Full} & [Misdirect] the letter .           & The pedestrian [misdirected] the out - of - town driver .    & F &F               \\ 
\textsc{Context}         &    [MASK] the letter .           & The pedestrian [MASK] the out - of - town driver .      & F & T                \\
\textsc{GuessedWord}         &    [Ignore] the letter . &
The pedestrian [ignored] the out - of - town driver .      & F & T\\
% \textsc{Advs Context} & [Misdirect] the letter to the out - of - town driver . & The pedestrian [misdirected] the out - of - town driver .&T&F\\
\hline
\textsc{Full}&[Kill] the engine .       &He [kills] the ball . &F&F\\
\textsc{Context}&[MASK] the engine       &He [MASK] the ball .&F&T \\
\textsc{GuessedWord} &[Hit] the engine .       &He [hits] the ball .&F&T \\
% \textsc{Advs Context}&[Kill] the engine and pass the ball .       &He [kills] the ball . &T&F\\
\hline

\textsc{Full}&[Kill] the engine .       &He [kills] the ball . &F&F\\
\textsc{Context}&[MASK] the engine       &He [MASK] the ball .&F&T \\
\textsc{GuessedWord} &[Hit] the engine .       &He [hits] the ball .&F&T \\

\hline
\textsc{Full} & His [treatment] of the race question is badly biased . & His [treatment] of space borrows from Italian architecture . & F & F\\
\textsc{Context} & His [MASK] of the race question is badly biased . & His [MASK] of space borrows from Italian architecture . & F & T \\
\textsc{GuessedWord} & His [understanding] of the race question is badly biased . & His [understanding] of space borrows from Italian architecture . & T & F \\
\hline
\textsc{Full} & I could see it in the [distance] .&	The [distance] from New York to Chicago .	& F & F \\
\textsc{Context} & I could see it in the [MASK] .&	The [MASK] from New York to Chicago .	& F & T \\
\textsc{GuessedWord} & I could see it in the [train] .&	The [train] from New York to Chicago .	& T & F \\

\bottomrule
\end{tabularx}
\caption{Example input of \textsc{Word}, \textsc{Context} and \textsc{Full} in WiC. The original WiC label for these examples is F. \textsc{GuessedWord} contains human-elicited target words that flip the label. Comparing \textsc{Context} and \textsc{GuessedWord} also shows \textsc{BERT}'s contextual bias in WiC as \textsc{BERT} is not sensitive to the target word change. \label{table:wic_exam_more}}
}
\end{table*}

% See Table~\ref{tab:errana_med} for an example where the context distracts the retrieval model to select the entry that is closer to the context than to the target word, but in fact knowing the target word alone is sufficient to retrieve the correct label. 

\section{Examples of the context bias in WiC\label{app:wic}}
See Table~\ref{table:wic_exam_more} for two examples where the model relies solely on the context to make the prediction.  

\section{Model configurations\label{app:model}}

ALL PCMs are from https://huggingface.co/. Model configurations are listed in Table~\ref{table:model}.\\

\begin{table*}
\scriptsize{
\begin{tabularx}{\textwidth}{p{0.1\textwidth}p{0.23\textwidth}p{0.32\textwidth}p{0.25\textwidth}} 
% {\scriptsize

\toprule
{\bf Model}&{\bf Variant name in Huggingface}&{\bf Parameters}&{\bf Pretraining corpus}\\
\midrule
     \textsc{BERT}& bert-base-uncased & 12-layer, 768-hidden, 12-heads, 110M parameters & Lowercased Wikipedia + BookCorpus\\
     \textsc{BERT}& bert-base-multilingual-uncased & 12-layer, 768-hidden, 12-heads, 110M parameters & Lowercased Wikipedia\\
     \textsc{RoBERTa}& roberta-base & 12-layer, 768-hidden, 12-heads, 110M parameters & Wikipedia + BookCorpus\\
     \textsc{PubMedBERT}&microsoft/BiomedNLP-PubMedBERT-base-uncased-abstract-fulltext & 12-layer, 768-hidden, 12-heads, 110M parameters & Lowercased abstracts from PubMed and full-text articles from PubMedCentral\\
\bottomrule
% }
\end{tabularx}}
\caption{Model details in our experiments\label{table:model}}
\end{table*}

\begin{figure}
\includegraphics[width=0.5\textwidth]{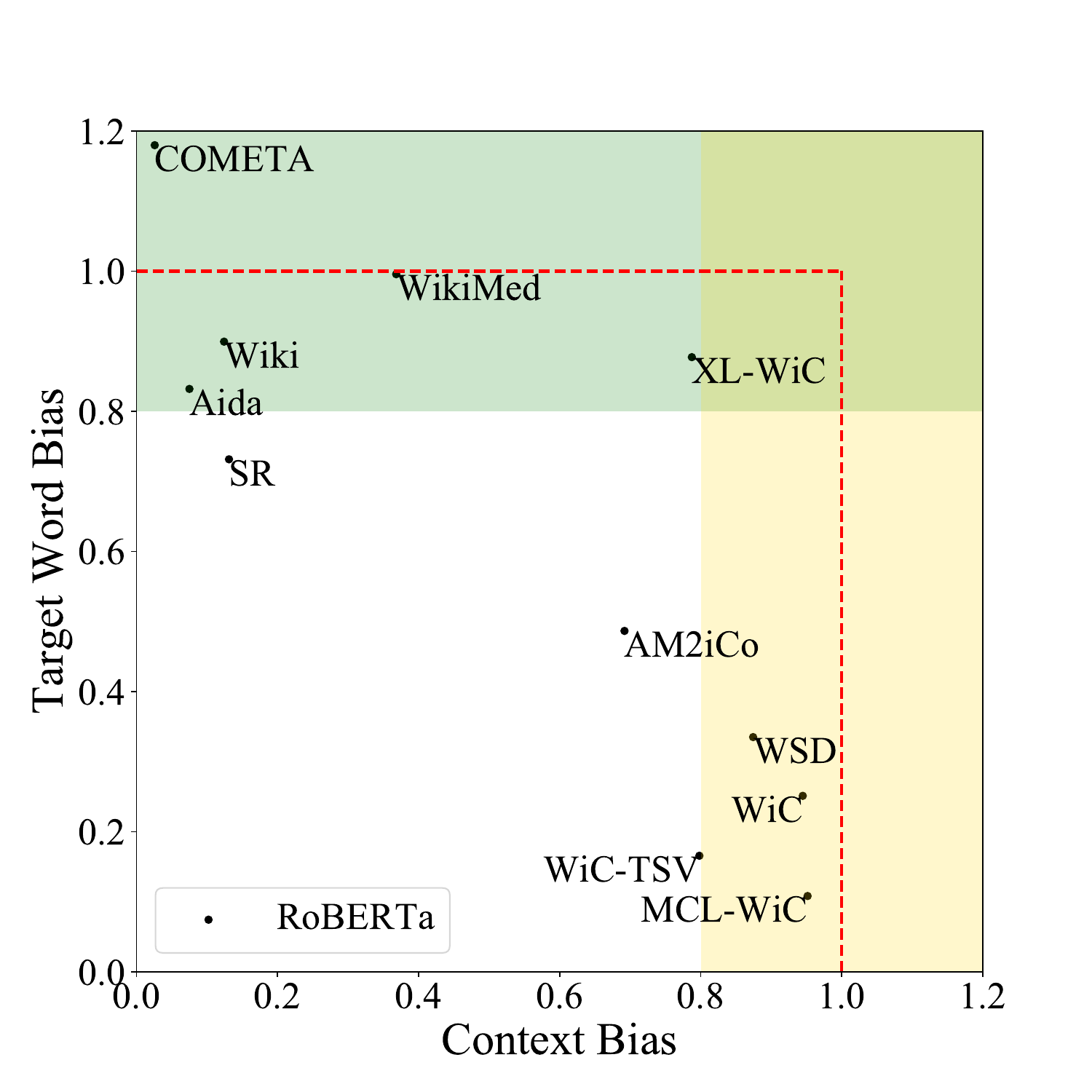}
\caption{Plotting context and target word biases when applying \textsc{RoBERTa} across popular context-aware lexical semantic datasets. The green shade and the yellow shade roughly indicate the areas for high target word bias and high context bias (0.8). The dashed red lines indicate 1.0 context (right) and 1.0 target word bias (top), implying the model only requires the target words alone or context alone in this dataset. 
 \label{fig:roberta_main}}
\end{figure}

\section{\textsc{RoBERTa} Performance\label{app:roberta} (\Cref{fig:roberta_main})}

% \Cref{fig:roberta_main} shows context and target word biases based on RoBERTa's performance across popular context-aware lexical semantic tasks. 

\section{Agreement in WiC-style tasks \label{sec:agreement}  (Table~\ref{table:agreement})}

\begin{table}[ht]%[t!]
\centering
\scriptsize{

\begin{tabular}{lccc}
\toprule 
& \amico &	XL-WiC&		WiC \\ 
\midrule
\textsc{Context}&94.0&88 & 76\\
\textsc{Full} &87.9&66 & 64\\

\bottomrule

\end{tabular}
\caption{Human agreement in \textsc{Context} and \textsc{Full} in WiC-style tasks\label{table:agreement}}}
\end{table}

\section{Annotation Guideline \label{sec:guideline}}

\begin{figure*}
\includegraphics[width=\textwidth]{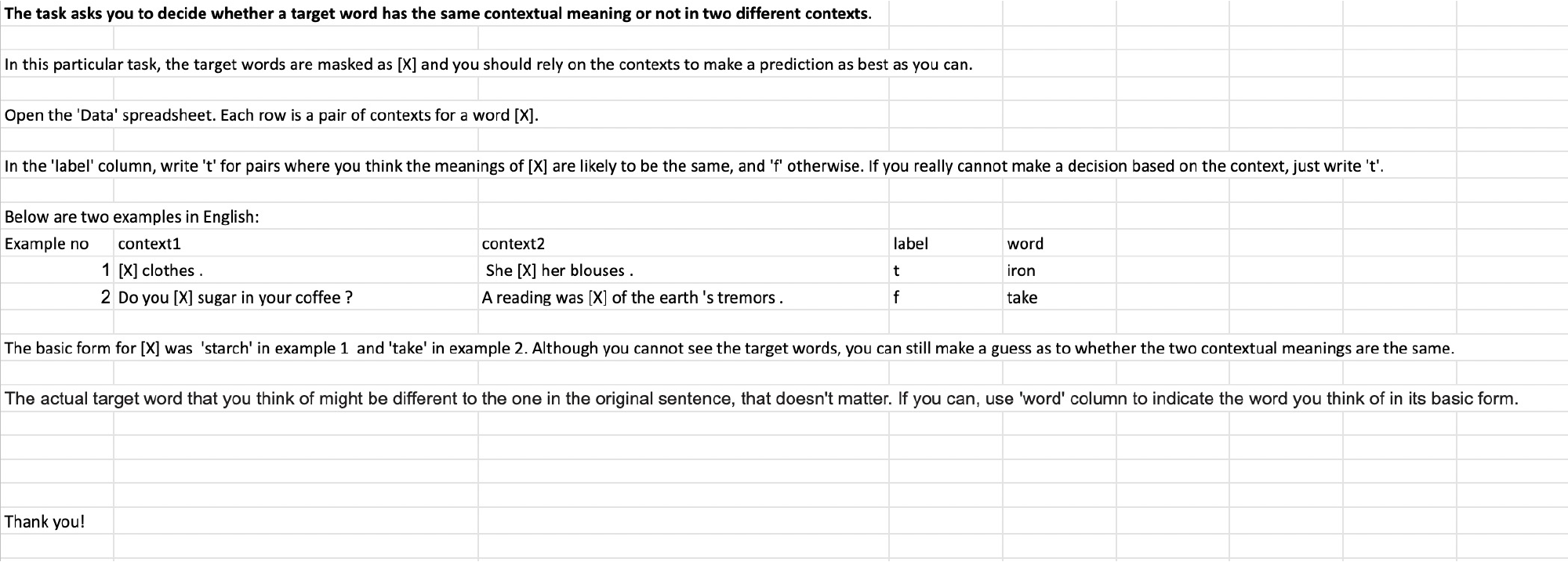}
\caption{An annotation guideline for conducting the \textsc{Context} baseline of humans in WiC.
 \label{fig:guideline}}
\end{figure*}

Figure~\ref{fig:guideline} shows an example annotation guideline for the \textsc{Context} experiment in WiC. 

% \section{\textsc{BERT} performance on probing baselines\label{app:bert_baselines} (\Cref{fig:baselines})}

% \begin{figure}

% \includegraphics[width=0.5\textwidth]{baslines_bert.jpeg}
% \caption{\textsc{BERT} performance on probing baselines across popular context-aware lexical semantic tasks. A small gap between \textsc{Full} and \textsc{Context}/\textsc{Word} baselines indicates strong bias on context/target word. For the retrieval-based tasks, we report @1 accuracy, and the \textsc{Label} and \textsc{Random} baselines are not visible as they are close to 0. 
%  \label{fig:baselines}}
% \end{figure}

\end{document}